\documentclass[twoside,11pt]{article}

\usepackage{amsmath,bm,paralist}
\usepackage{algorithm,algorithmic}
\usepackage{jmlr2e}
\usepackage[colorlinks,
            linkcolor=blue,
            citecolor=blue,
            urlcolor=magenta,
            linktocpage,
            plainpages=false]{hyperref}

\makeatletter
\AtBeginDocument{\Hy@breaklinkstrue}
\makeatother

\newtheorem{thm}{Theorem}

\newtheorem{ass}{Assumption}
\newtheorem{cor}[thm]{Corollary}
\newtheorem{defit}{Definition}

\def \y {\mathbf{y}}
\def \E {\mathrm{E}}
\def \x {\mathbf{x}}

\def \u {\mathbf{u}}

\def \w {\mathbf{w}}
\def \R {\mathbb{R}}

\def \P {\mathcal{P}}

\def \ph {\widehat{p}}

\def \I {\mathcal{I}}

\def \S {\mathcal{S}}
\DeclareMathOperator*{\Reg}{Regret}
\DeclareMathOperator*{\WAReg}{WA-Regret}
\DeclareMathOperator*{\SAReg}{SA-Regret}
\DeclareMathOperator*{\DReg}{D-Regret}
\DeclareMathOperator*{\poly}{poly}
\DeclareMathOperator*{\argmin}{argmin}

\makeatletter
\newcounter{ALC@tempcntr}
\newcommand{\LCOMMENT}[1]{%
    \setcounter{ALC@tempcntr}{\arabic{ALC@rem}}
    \setcounter{ALC@rem}{1}
    \item \{#1\}
    \setcounter{ALC@rem}{\arabic{ALC@tempcntr}}
}%
\makeatother

\begin{document}

\title{Dynamic Regret of Strongly Adaptive Methods}

\author{\name Lijun Zhang \email zhanglj@lamda.nju.edu.cn\\
       \addr National Key Laboratory for Novel Software Technology\\
       Nanjing University, Nanjing 210023, China
       \AND
       \name Tianbao Yang \email tianbao-yang@uiowa.edu\\
       \addr Department of Computer Science\\
       the University of Iowa, Iowa City, IA 52242, USA
       \AND     
       \name Rong Jin \email jinrong.jr@alibaba-inc.com\\
       \addr Alibaba Group, Seattle, USA
       \AND
       \name Zhi-Hua Zhou \email zhouzh@lamda.nju.edu.cn\\
       \addr National Key Laboratory for Novel Software Technology\\
       Nanjing University, Nanjing 210023, China}

\editor{}

\maketitle

\begin{abstract}
To cope with changing environments, recent developments in online learning have introduced the concepts of \emph{adaptive} regret and \emph{dynamic} regret independently. In this paper, we illustrate an intrinsic connection between these two concepts by showing that the dynamic regret can be expressed in terms of the adaptive regret and the functional variation. This observation implies that strongly adaptive algorithms can be directly leveraged to minimize the dynamic regret. As a result, we present a series of strongly adaptive algorithms that have small dynamic regrets for convex functions, exponentially concave functions, and strongly convex functions, respectively. To the best of our knowledge, this is the \emph{first} time that exponential concavity is utilized to upper bound the dynamic regret. Moreover, all of those adaptive algorithms do not need any prior knowledge of the functional variation, which is a significant advantage over previous specialized methods for minimizing dynamic regret.
\end{abstract}


\section{Introduction}
Online convex optimization is a powerful paradigm for sequential decision making \citep{zinkevich-2003-online}. It can be viewed as a game between a learner and an adversary: In the $t$-th round, the learner selects a decision $\w_t \in \Omega$, simultaneously the adversary chooses a function $f_t(\cdot): \Omega \mapsto \R$, and then the learner suffers an instantaneous loss $f_t(\w_t)$.  This study focuses on the full-information setting, where the learner can query the value and gradient of $f_t$  \citep{bianchi-2006-prediction}.  The goal of the learner is to minimize the cumulative loss over $T$ periods . The standard performance measure is \emph{regret}, which is the difference between the loss incurred by the learner and that of the best fixed decision in hindsight, i.e.,
\[
\Reg(T)=\sum_{t=1}^T f_t(\w_t) - \min_{\w \in \Omega} \sum_{t=1}^T f_t(\w).
\]

The above regret is typically referred to as \emph{static} regret in the sense that the comparator is time-invariant. The rationale behind this evaluation metric is that one of the decision in $\Omega$ is reasonably good over the $T$ rounds. However, when  the underlying distribution of loss functions
changes, the static regret may be too optimistic and fails to capture the hardness of the problem. 

To address this limitation, new forms of performance measure, including \emph{adaptive} regret \citep{Adaptive:Hazan,Hazan:2009:ELA} and \emph{dynamic} regret \citep{zinkevich-2003-online,Dynamic:ICML:13}, were proposed and received significant interest  recently.  Following the terminology of \citet{Adaptive:ICML:15}, we define the strongly adaptive regret as the maximum static regret over intervals of length $\tau$, i.e.,
\begin{equation} \label{eqn:strong:adaptive}
\begin{split}
\SAReg(T,\tau) = \max_{[s, s+\tau -1] \subseteq [T]} \left(\sum_{t=s}^{s+\tau -1}  f_t(\w_t)  - \min_{\w \in \Omega} \sum_{t=s}^{s+\tau -1} f_t(\w) \right).
\end{split}
\end{equation}
Minimizing the adaptive regret enforces the learner to have a small static regret over any interval of length $\tau$. Since the best decision for different intervals could be different, the learner is essentially competing with a changing comparator. 

A parallel line of research introduces the concept of dynamic regret, where the cumulative loss of the learner is compared against a comparator sequence $\u_1, \ldots, \u_T \in \Omega$, i.e.,
\begin{equation} \label{eqn:dynamic:1}
\DReg(\u_1,\ldots,\u_T) = \sum_{t=1}^T f_t(\w_t)  - \sum_{t=1}^T f_t(\u_t).
\end{equation}
It is well-known that in the worst case, a sublinear dynamic regret is impossible unless we impose some regularities on the comparator sequence or the function sequence \citep{Oinline:Dynamic:Comp}. A representative example is the functional variation  defined below
\begin{equation} \label{eqn:func:var}
V_T = \sum_{t=2}^T \max_{\w \in \Omega} |f_t(\w) - f_{t-1}(\w)|.
\end{equation}
\citet{Non-Stationary} have  proved that as long as $V_T$ is sublinear in $T$, there exists an algorithm that achieves a sublinear dynamic regret. Furthermore,  a general restarting procedure is developed, and it enjoys $O(T^{2/3}V_T^{1/3})$ and $O(\log T \sqrt{T V_T})$ rates for convex functions and strongly convex functions, respectively. 
However, the restarting procedure can only be applied when an upper bound of $V_T$ is known beforehand, thus limiting its application in practice.

While both the adaptive and dynamic regrets aim at coping with changing environments, little is known about their relationship. This paper makes a step towards understanding their connections. Specifically, we show that the strongly adaptive regret in (\ref{eqn:strong:adaptive}), together with the functional variation, can be used to upper bound the dynamic regret in (\ref{eqn:dynamic:1}). Thus, an algorithm with a small strongly adaptive regret is automatically equipped with a tight dynamic regret. As a result, we obtain a series of algorithms for minimizing the dynamic regret that do not need any prior knowledge of the functional variation.  The main contributions of this work are summarized below.
\begin{compactitem}
\item We provide a general theorem that  upper bounds the dynamic regret in terms of the strongly adaptive regret and the functional variation.
\item For convex functions, we show that the strongly adaptive algorithm of \citet{Improved:Strongly:Adaptive} has a dynamic regret of  $O(T^{2/3} V_T^{1/3} \log^{1/3} T)$, which matches the minimax rate \citep{Non-Stationary},  up to a polylogarithmic factor.
\item For exponentially concave functions, we propose a strongly adaptive algorithm that allows us to control the tradeoff between the adaptive regret and the computational cost explicitly. Then, we demonstrate that its dynamic regret is $O(d \sqrt{T V_T \log T})$, where $d$ is the dimensionality. To the best of our knowledge, this is the \emph{first} time that exponential concavity is utilized in the analysis of dynamic regret.
\item For strongly convex functions,  our proposed algorithm can also be applied and yields a dynamic regret of $O(\sqrt{T V_T \log T})$, which is also minimax optimal up to a polylogarithmic factor.
\end{compactitem}

\section{Related Work}
We give a brief introduction to previous work on static, adaptive, and dynamic regrets in the context of online convex optimization.
\subsection{Static Regret}
The majority of studies in online learning are focused on static regret \citep{Shalev:Primal:Dual,Sparse:Online,Online:suvery,ICML:13:Zhang:Sparse}. For general convex functions, the classical online gradient descent achieves $O(\sqrt{T})$ and $O(\log T)$ regret bounds for convex and strongly convex functions, respectively \citep{zinkevich-2003-online,ML:Hazan:2007,ICML_Pegasos}.  Both the $O(\sqrt{T})$ and $O(\log T)$ rates are known to be minimax optimal~\citep{Minimax:Regret}. When functions are exponentially concave, a different algorithm, named online Newton step, is developed and enjoys an $O(d \log T)$ regret bound, where $d$ is the dimensionality \citep{ML:Hazan:2007}.

\subsection{Adaptive Regret}
The concept of adaptive regret is  introduced by \citet{Adaptive:Hazan}, and later strengthened  by \citet{Adaptive:ICML:15}. Specifically,   \citet{Adaptive:Hazan} introduce the weakly adaptive regret
\[
\begin{split}
\WAReg(T)=  \max_{[s, q] \subseteq [T]} \left(\sum_{t=s}^{q} f_t(\w_t) - \min_{\w \in \Omega} \sum_{t=s}^{q} f_t(\w)\right).
\end{split}
\]
To minimize the adaptive regret, \citet{Adaptive:Hazan} have developed two meta-algorithms: an efficient algorithm with $O(\log T)$ computational complexity per iteration and an inefficient one with $O(T)$ computational complexity per iteration. These meta-algorithms use an existing online method (that was possibly designed to have small static regret) as a subroutine.\footnote{For brevity, we ignored the factor of subroutine in the statements of computational complexities. The $O(\cdot)$ computational complexity should be interpreted as $O(\cdot) \times s$ space complexity and $O(\cdot) \times t$ time complexity, where $s$ and $t$ are  space  and time complexities of the subroutine per iteration, respectively.} For  convex functions, the efficient and inefficient meta-algorithms have   $O(\sqrt{T \log^3 T})$ and $O(\sqrt{T \log T)}$ regret bounds, respectively. For exponentially concave functions, those rates are improved to  $O(d \log^2 T)$ and $O(d \log T)$, respectively. We can see that the price paid for the adaptivity is very small: The rates of weakly adaptive regret differ from those of static regret only by logarithmic factors.

A major limitation of weakly adaptive regret is that it does not respect short intervals well. Taking convex functions as an example, the $O(\sqrt{T \log^3 T})$ and $O(\sqrt{T \log T)}$ bounds are meaningless for intervals of length $O(\sqrt{T})$. To overcome this limitation, \citet{Adaptive:ICML:15} proposed the strongly adaptive regret  $\SAReg(T,\tau)$ which takes the length of the interval $\tau$ as a parameter, as indicated in (\ref{eqn:strong:adaptive}). From the definitions, we have $\SAReg(T,\tau) \leq \WAReg(T)$, but it does not mean  the notation of weakly adaptive regret is stronger, because an upper bound for $\WAReg(T)$ could be very loose for $\SAReg(T,\tau)$ when $\tau$ is small.

If the strongly adaptive regret is small for all $\tau <T$, we can guarantee the learner has a small regret over any interval of any length. In particular, \citet{Adaptive:ICML:15} introduced the following definition.
\begin{defit} \label{def:strongly:adaptive}
Let $R(\tau)$ be the minimax static regret bound of the learning problem over $\tau$ periods. An algorithm is \emph{strongly adaptive}, if
\[
\SAReg(T,\tau)=O(\poly(\log T) \cdot R(\tau)), \ \forall \tau.
\]
\end{defit}
It is easy to verify that the meta-algorithms of \citet{Adaptive:Hazan} are strongly adaptive for exponentially concave functions,\footnote{That is because (i) $\SAReg(T,\tau) \leq \WAReg(T)$, and (ii) there is a $\poly(\log T)$ factor in the definition of strong adaptivity. } but not for convex functions. Thus, \citet{Adaptive:ICML:15} developed a new meta-algorithm that satisfies $\SAReg(T,\tau)=O( \sqrt{\tau} \log T )$ for convex functions, and thus is strongly adaptive. The algorithm is also efficient and the computational complexity per iteration is $O(\log T)$. Later, the strongly adaptive regret of convex functions was improved to $O( \sqrt{\tau \log T} )$ by \citet{Improved:Strongly:Adaptive}, and the computational complexity remains $O(\log T)$ per iteration. All the previously mentioned algorithms for minimizing adaptive regret need to query the gradient of the loss function at least $O(\log t)$ times in the $t$-th iteration. In a recent study, \citet{Adaptive:One:Gradient} demonstrate that the number of gradient evaluations per iteration can be reduced to $1$ by introducing the surrogate loss.

\subsection{Dynamic Regret} \label{sec:dynamic}
In a seminal work, \citet{zinkevich-2003-online} proposed to use the \emph{path-length} defined as
\[
\P(\u_1, \ldots, \u_T)=\sum_{t=2}^T \|\u_t - \u_{t-1}\|_2
\]
to upper bound the dynamic regret, where $\u_1, \ldots, \u_T \in \Omega$ is a comparator sequence. Specifically, \citet{zinkevich-2003-online} proved that for any sequence of convex functions, the dynamic regret of online gradient descent can be upper bounded by $O(\sqrt{T} \P(\u_1, \ldots, \u_T))$. Another regularity of the comparator sequence, which is similar to the path-length, is defined as
\[
 \P'(\u_1, \ldots, \u_T)=\sum_{t=2}^T \|\u_t - \Phi_{t} (\u_{t-1})\|_2
\]
where $\Phi_t (\cdot)$ is a dynamic model that predicts a reference point for the $t$-th round.  \citet{Dynamic:ICML:13} developed a novel algorithm named dynamic mirror descent and proved that its dynamic regret is on the order of $\sqrt{T} \P'(\u_1, \ldots, \u_T)$. The advantage of $\P'(\u_1, \ldots, \u_T)$  is that when the comparator sequence follows the dynamical model closely, it can be much smaller than the path-length $\P(\u_1, \ldots, \u_T)$.

Let $\w_t^* \in \argmin_{\w \in \Omega} f_t(\w)$ be a minimizer of $f_t(\cdot)$. For any sequence of $\u_1, \ldots, \u_T \in \Omega$, we have
\[
\begin{split}
& \DReg(\u_1,\ldots,\u_T) =\sum_{t=1}^T f_t(\w_t)  - \sum_{t=1}^T f_t(\u_t) \\
\leq & \DReg(\w_1^*,\ldots,\w_T^*) = \sum_{t=1}^T f_t(\w_t)  - \sum_{t=1}^T \min_{\w \in \Omega} f_t(\w).
\end{split}
\]
Thus, $\DReg(\w_1^*,\ldots,\w_T^*)$ can be treated as the worst case of the dynamic regret, and there are many works that were devoted to minimizing $\DReg(\w_1^*,\ldots,\w_T^*)$ \citep{Oinline:Dynamic:Comp,Dynamic:Strongly,Dynamic:2016,Dynamic:Regret:Squared}.

When a prior knowledge of $\P(\w_1^*, \ldots, \w_T^*)$ is available, $\DReg(\w_1^*,\ldots,\w_T^*)$ can be upper bounded by $O(\sqrt{T \P(\w_1^*, \ldots, \w_T^*)})$ \citep{Dynamic:2016}. If all the functions are strongly convex and smooth, the upper bound can be improved to $O(\P(\w_1^*, \ldots, \w_T^*))$ \citep{Dynamic:Strongly}. The $O(\P(\w_1^*, \ldots, \w_T^*))$ rate is also achievable when all the functions are convex and smooth, and all the minimizers $\w_t^*$'s lie in the interior of $\Omega$ \citep{Dynamic:2016}. In a recent study, \citet{Dynamic:Regret:Squared} introduced a new regularity---\emph{squared} path-length
\[
\S(\w_1^*, \ldots, \w_T^*)=\sum_{t=2}^T \|\w_t^* - \w_{t-1}^*\|_2^2
\]
which could be much smaller than the path-length $\P(\w_1^*, \ldots, \w_T^*)$ when the difference between successive  minimizers is small. \citet{Dynamic:Regret:Squared} developed a novel algorithm named online multiple gradient descent, and proved that $\DReg(\w_1^*,\ldots,\w_T^*)$ is on the order of $\min(\P(\w_1^*, \ldots, \w_T^*),\S(\w_1^*, \ldots, \w_T^*))$ for (semi-) strongly convex and smooth functions.

\paragraph{Discussions} Although closely related, adaptive regret and dynamic regret are studied independently and there are few discussions of their relationships. In the literature, dynamic regret is also referred to as tracking regret or shifting regret \citep{LITTLESTONE1994212,Herbster1998,Herbster:2001:TBL}. In the setting of ``prediction with expert advice'', \citet{Adamskiy2012} have shown that the tracking regret can be derived from the adaptive regret.  In the setting of ``online linear optimization in the simplex'', \citet{Fixed:Share:NIPS12} introduced a generalized notion of shifting regret which unifies  adaptive regret and shifting regret. Different from previous work, this paper considers the  setting of online convex optimization, and illustrates that the dynamic regret can be upper bounded by the adaptive regret and the functional variation.
\section{A Unified Adaptive Algorithm} \label{sec:adaptive:alg}
In this section, we introduce a unified approach for minimizing the adaptive regret of exponentially concave functions, as well as strongly convex functions.

\subsection{Motivation}
We first provide the definition of exponentially concave (abbr.~exp-concave) functions  \citep{bianchi-2006-prediction}.
\begin{defit} \label{def:exp} A function $f(\cdot): \Omega \mapsto \R$ is $\alpha$-exp-concave if $\exp(-\alpha f(\cdot))$ is concave over $\Omega$.
\end{defit}

For exp-concave functions, \citet{Adaptive:Hazan} have developed two meta-algorithms that take the online Newton step as its subroutine, and proved the following properties.
\begin{compactitem}
  \item The inefficient one has $O(T)$ computational complexity per iteration, and its adaptive regret is $O(d \log T)$.
  \item The efficient one has $O(\log T)$ computational complexity per iteration, and its adaptive regret is $O(d \log^2 T)$.
\end{compactitem}
As can be seen, there is a tradeoff between the computational complexity and the adaptive regret: A lighter computation incurs a looser bound and a tighter bound requires a  higher computation.  Our goal is to develop  a unified approach, that allows us to trade effectiveness for efficiency explicitly.
\subsection{Improved Following the Leading History (IFLH)}
Let $E$ be an online learning algorithm that is designed to minimize the static regret of exp-concave functions or strongly convex functions, e.g.,  online Newton step \citep{ML:Hazan:2007} or online gradient descent \citep{zinkevich-2003-online}. Similar to the approach of following the leading history (FLH) \citep{Adaptive:Hazan}, at any time $t$, we will instantiate an expert by applying the online learning algorithm $E$ to the sequence of loss functions $f_t,f_{t+1},\ldots$, and utilize the strategy of learning from expert advice to combine solutions of different experts \citep{Herbster1998}. Our method is named as improved following the leading history (IFLH), and is summarized in Algorithm~\ref{alg:1}.

Let $E^t$ be the expert that starts to work at time $t$. To control the computational complexity, we will associate an ending time $e^t$ for each $E^t$.  The expert $E^t$ is alive during the period $[t, e^t-1]$. In each round $t$, we maintain a working set of experts $\S_t$, which contains all the alive experts, and assign a probability $p_t^j$ for each $E^j \in S_t$. In Steps~6 and 7, we remove all the experts whose ending times are no larger than $t$. Since the number of alive experts has changed, we need to update the probability assigned to them, which is performed in Steps~12 to 14. In Steps~15 and 16, we add a new expert $E^t$ to $\S_t$, calculate its ending time according to Definition~\ref{def:ending} introduced below, and set $p_t^t = \frac{1}{t}$. It is easy to verify $\sum_{E^j \in \S_t} p_t^j=1$. Let $\w_t^j$ be the output of $E^j$ at the $t$-th round, where $t \geq j$. In Step~17, we submit the weighted average of $\w_t^j$ with coefficient $p_t^j$ as the output $\w_t$, and suffer the loss $f_t(\w_t)$. From Steps~18 to 25, we use the exponential weighting scheme to update the weight for each expert $E^j$ based on its loss $f_t(\w_{t}^j)$. In Step~21, we pass the loss function to all the alive experts such that they can update their predictions for the next round.

\begin{algorithm}[t]
    \caption{Improved Following the Leading History (IFLH)} \label{alg:1}
    \begin{algorithmic}[1]
    \STATE {\bf Input:} An integer $K$
    \STATE Initialize $\S_0 = \emptyset$.
    \FOR{$t = 1, \ldots, T$}
        \STATE Set $Z_t = 0$
        \LCOMMENT{Remove some existing experts}
        \FOR{$E^j \in \S_{t-1}$} \label{step:1}
            \IF{$e^j \leq t$}\label{step:2}
                \STATE Update $\S_{t-1} \leftarrow \S_{t-1} \setminus \{E^j\}$ \label{step:3}
            \ELSE \label{step:4}
                \STATE Set $Z_t = Z_t + \ph_t^j$ \label{step:5}
            \ENDIF \label{step:6}
        \ENDFOR
       \LCOMMENT{Normalize the probability}
        \FOR{$E^j \in \S_{t-1}$} \label{step:7}
            \STATE Set $p_t^j = \frac{\ph_t^j}{Z_t}\left(1 - \frac{1}{t}\right)$ \label{step:8}
        \ENDFOR \label{step:9}
        \LCOMMENT{Add a new expert $E^t$}
        \STATE Set $\S_t = \S_{t-1} \cup \{E^t\}$  \label{step:10}
        \STATE  Compute the ending time $e^t=\E_K(t)$ according to Definition~\ref{def:ending} and set $p_t^t = \frac{1}{t}$ \label{step:11}
        \LCOMMENT{Compute the final predicted model}
        \STATE Submit the solution
        \[
        \w_t = \sum_{E^j  \in \S_t} p_t^j \w_t^j
        \]
         and suffer loss $f_t(\w_t)$  \label{step:12}
        \LCOMMENT{Update weights and expert}
        \STATE Set $Z_{t+1} = 0$ \label{step:13}
        \FOR{$E^j \in \S_t$}
            \STATE Compute $p_{t+1}^j = p_t^j \exp(-\alpha f_t(\w_{t}^j))$ and $Z_{t+1} = Z_{t+1} + p_{t+1}^j$
            \STATE Pass the function $f_t(\cdot)$ to $E^j$ \label{step:14}
        \ENDFOR
        \FOR{$E^j \in \S_t$}
            \STATE Set $\ph_{t+1}^j = \frac{p_{t+1}^j}{Z_{t+1}}$
        \ENDFOR \label{step:15}
    \ENDFOR
    \end{algorithmic}
\end{algorithm}
The difference between our IFLH and the original FLH is how to decide the ending time $e^t$ of  expert $E^t$. In this paper, we propose the following base-$K$ ending time.
\begin{defit}[Base-$K$ Ending Time] \label{def:ending} Let $K$ be an integer, and the representation of $t$ in the base-$K$ number system as
\[
t= \sum_{\tau \geq 0} \beta_\tau K^\tau
\]
where $0 \leq \beta_\tau <K$, for all $\tau \geq 0$. Let $k$ be the smallest integer such that $\beta_k > 0$, i.e., $k = \min\{\tau:\beta_\tau > 0\}$. Then, the base-$K$ ending time of $t$ is defined as
\[
\E_K(t)=\sum_{\tau \geq k+1} \beta_\tau K^\tau + K^{k+1}.
\]
In other words, the ending time is the number represented by the new sequence obtained by setting the first nonzero element in the sequence $\beta_0,\beta_1,\ldots$ to be $0$ and adding $1$ to the element after it.
\end{defit}
 Let's take the decimal system as an example (i.e., $K=10$). Then,
 \[
 \begin{split}
&E_{10}(1)=E_{10}(2)=\cdots =E_{10}(9)=10,\\
&E_{10}(11)=E_{10}(12)=\cdots=E_{10}(19)=20,\\
&E_{10}(10)=E_{10}(20)=\cdots=E_{10}(90)=100.\\
\end{split}
\]

\subsection{Theoretical Guarantees}
When the base-$K$ ending time is used in Algorithm~\ref{alg:1}, we have the following properties.
\begin{lemma} \label{lem:ending}
Suppose we use the base-$K$ ending time in Algorithm~\ref{alg:1}.
\begin{compactenum}
\item For any $t \geq 1$, we have
\[
|\S_t| \leq \left(\lfloor \log_K t \rfloor+1 \right) (K-1)=O\left( \frac{K \log t}{\log K} \right).
\]
\item  For any interval $I = [r, s] \subseteq [T]$, we can always find $m$ segments $I_j = [t_j, e^{t_j}-1], \ j \in [m]$
with $m \leq \lceil\log_K (s-r+1)\rceil +1$, such that $t_1=r, \ e^{t_j}=t_{j+1},  \ j \in [m-1], \textrm{ and } e^{t_m} > s$.
\end{compactenum}
\end{lemma}
The first part of Lemma~\ref{lem:ending} implies that the size of $\S_t$ is $O(K \log t/\log K)$. An example of $\S_t$ in the decimal system is given below.
 \[
 \S_{486}=\left\{ \begin{split}
 &481, \ 482, \ \ldots, \ 486, \\
   &410, \ 420, \ \ldots, \ 480,\\
   &100, \ 200, \ \ldots, \ 400
 \end{split} \right\} .
 \]
The second part of Lemma~\ref{lem:ending} implies that for any interval $I=[r, s]$, we can find $O(\log s/\log K)$ experts such that their survival periods cover $I$. Again, we present an example in the decimal system: The interval $[111, 832]$ can be covered by
\[
[111, 119], \ [120, 199], \textrm{ and } [200, 999]
\]
which are the survival periods of experts $E^{111}$, $E^{120}$, and $E^{200}$,  respectively. Recall that $E_{10}(111)=120$, $E_{10}(120)=200$, and $E_{10}(200)=1000$.

We note that a similar strategy for deciding the ending time was proposed by \citet{Track_Large_Expert} in the study of ``prediction with expert advice''. The main difference is that their strategy  is built upon base-$2$ number system and introduces an additional parameter $g$ to compromise between the computational complexity and the regret, in contrast our method relies on base-$K$ number system and uses $K$ to control the tradeoff. Lemma~2 of \citet{Track_Large_Expert} indicates an $O(g \log t )$ bound on the number of alive experts, which is worse than our $O(K \log t/\log K)$ bound by a logarithmic factor.

To present adaptive regret bounds, we introduce the following common assumption.
\begin{ass}\label{ass:1} Both the gradient and the domain are bounded.
\begin{compactitem}
\item The gradients of all the online functions are bounded by $G$, i.e., $\max_{\w \in \Omega}\|\nabla f_t(\w)\| \leq G$ for all $f_t$.
\item The diameter of the domain $\Omega$ is bounded by $B$, i.e., $\max_{\w, \w' \in \Omega}\|\w -\w'\| \leq B$.
\end{compactitem}
\end{ass}
Based on Lemma~\ref{lem:ending}, we have the following theorem regarding the adaptive regret of exp-concave functions.
\begin{thm} \label{thm:adaptive} Suppose Assumption~\ref{ass:1} holds, $\Omega \subset \R^d$, and all the functions are $\alpha$-exp-concave. If  online Newton step is used as the subroutine in Algorithm~\ref{alg:1}, we have
\[
\begin{split}
\sum_{t=r}^s f_t(\w_t) - \min\limits_{\w \in \Omega} \sum_{t=r}^s f_t(\w)\leq  \left(\frac{(5d+1) m + 2}{\alpha} + 5d mGB \right) \log T
\end{split}
\]
where $[r,s] \subseteq [T]$ and $m \leq \lceil\log_K (s-r+1)\rceil +1$. Thus,
\[
\begin{split}
\SAReg(T,\tau)\leq \left(\frac{(5d+1) \bar{m} + 2}{\alpha} + 5d \bar{m} GB \right) \log T = O\left( \frac{d \log^2 T }{\log K}\right)
 \end{split}
\]
where $\bar{m}=\lceil\log_K \tau \rceil +1$.
\end{thm}
From Lemma~\ref{lem:ending} and Theorem~\ref{thm:adaptive}, we observe that the adaptive regret is a decreasing function of $K$, while the computational cost is an increasing function of $K$. Thus, we can control the tradeoff by tuning the value of $K$.
Specifically, Lemma~\ref{lem:ending} indicates the proposed algorithm has
\[
\left(\lfloor \log_K T \rfloor+1 \right) (K-1)=O\left( \frac{K \log T}{\log K} \right)
\]
computational complexity per iteration. On the other hand, Theorem~\ref{thm:adaptive} implies that for $\alpha$-exp-concave functions that satisfy Assumption~\ref{ass:1}, the strongly adaptive regret of Algorithm~\ref{alg:1} is
 \[
\left(\frac{(5d+1) \bar{m} + 2}{\alpha} + 5d \bar{m} GB \right) \log T = O\left(  \frac{d \log^2 T}{\log K}\right)
\]
where $d$ is the dimensionality and $\bar{m}= \lceil\log_K (\tau)\rceil +1$.

We list several choices of $K$ and the resulting theoretical guarantees in Table~\ref{sample-table}, and have the following observations.
\begin{compactitem}
\item When $K=2$, we recover the guarantee of the  efficient algorithm of \citet{Adaptive:Hazan}, and when $K=T$, we obtain the inefficient one.
  \item By setting $K=\lceil T^{1/\gamma}\rceil$ where $\gamma>1$ is a small constant, such as $10$, the strongly adaptive regret can be viewed as $O(d \log T)$, and at the same time,  the computational complexity is also very low for a large range of $T$. 
\end{compactitem}

\begin{table}[t]
\caption{Efficiency  and Effectiveness Tradeoff} \label{sample-table}
\begin{center}
\begin{tabular}{lll}
$K$  &Complexity  & Adaptive Regret\\
\hline \\
$2$         & $O(\log T)$ & $O(d \log^2 T)$ \\
$\lceil T^{1/\gamma}\rceil$   & $O(\gamma T^{1/\gamma} )$ & $O(\gamma d \log T)$ \\
$T$             & $O(T)$ & $O(d \log T)$ \\
\end{tabular}
\end{center}
\end{table}

Next, we consider strongly convex functions.
\begin{defit} A function $f(\cdot): \Omega \mapsto \R$ is $\lambda$-strongly convex if
\[
f(\y) \geq f(\x) + \left \langle \nabla f(\x), \y -\x \right \rangle + \frac{\lambda}{2} \|\y -\x \|_2^2,  \  \forall \x, \y \in \Omega.
\]
\end{defit}
It is easy to verify that strongly convex functions with bounded gradients are also exp-concave \citep{ML:Hazan:2007}.
\begin{lemma} \label{lem:strongly} Suppose $f(\cdot): \Omega \mapsto \R$ is $\lambda$-strongly convex and $\|\nabla f(\w)\| \leq G$ for all $\w \in \Omega$. Then, $f(\cdot)$ is $\frac{\lambda}{G^2}$-exp-concave.
\end{lemma}
According to the above lemma, we still use Algorithm~\ref{alg:1} as the meta-algorithm, but choose online gradient descent as the subroutine. In this way, the adaptive regret does not depend on the dimensionality $d$.

\begin{thm} \label{thm:strong:convex} Suppose Assumption~\ref{ass:1} holds, and  all the functions are $\lambda$-strongly convex. If online gradient descent is used as the subroutine in Algorithm~\ref{alg:1}, we have
\[
\begin{split}
 \sum_{t=r}^s f_t(\w_t) - \min\limits_{\w \in \Omega} \sum_{t=r}^s f_t(\w)
 \leq \frac{G^2}{2\lambda} \big(m+ (3 m +4)  \log T\big )
\end{split}
\]
where $[r,s] \subseteq [T]$ and $m \leq \lceil\log_K (s-r+1)\rceil +1$. Thus
\[
\begin{split}
\SAReg(T,\tau)\leq \frac{G^2}{2\lambda} \big(\bar{m}+ (3\bar{m} +4)  \log T \big)= O\left( \frac{\log^2 T }{\log K}\right)
 \end{split}
\]
where $\bar{m}=\lceil\log_K \tau \rceil +1$.
\end{thm}

\section{From Adaptive to Dynamic}
In this section, we first introduce a general theorem that bounds the dynamic regret by the adaptive regret, and then derive specific regret bounds  for convex functions, exponentially concave functions, and strongly convex functions.

\subsection{Adaptive-to-Dynamic Conversion}
Let $\I_1=[s_1, q_1], \I_2 = [s_2, q_2], \ldots, \I_k=[s_k, q_k]$ be a partition of $[1,T]$. That is, they are successive intervals such that
\begin{equation} \label{eqn:intervel}
\begin{split}
s_1=1, \ q_i +1 = s_{i+1},  \  i \in [k-1], \textrm{ and }   \ q_k=T.
\end{split}
\end{equation}
Define the local functional variation of the $i$-th interval as
\[
V_T(i) = \sum_{t=s_i+1}^{q_i} \max_{\w \in \Omega} |f_t(\w) - f_{t-1}(\w)|
\]
and it is obvious that $\sum_{i=1}^k V_T(i) \leq V_T$.\footnote{Note that in certain cases, the sum of local functional variation $\sum_{i=1}^k V_T(i)$ can be much smaller than the total functional variation $V_T$. For example, when the sequence of functions only changes $k$ times,  we can construct the intervals based on the changing rounds such that $\sum_{i=1}^k V_T(i)=0$.}
Then, we have the following theorem for bounding the dynamic regret in terms of the strongly adaptive regret and the functional variation.
\begin{thm} \label{thm:1} Let $\w_t^* \in \argmin_{\w \in \Omega} f_t(\w)$.  For all integer $k \in [T]$, we have
\[
\begin{split}
\DReg(\w_1^*,\ldots,\w_T^*) \leq  \min_{\I_1,\ldots,\I_k} \sum_{i=1}^k \big( \SAReg(T,|\I_i|) + 2 |\I_i| \cdot V_T(i) \big)
\end{split}
\]
where the minimization is taken over any sequence of intervals that satisfy (\ref{eqn:intervel}).
\end{thm}
The above theorem is analogous to Proposition 2 of \citet{Non-Stationary}, which provides an upper bound for a special choice of the interval sequence. The main difference is that there is a minimization operation in our bound, which allows us to get rid of the issue of parameter selection. For a specific type of problems, we can plug in the corresponding upper bound of strongly adaptive regret, and then choose any sequence of intervals to obtain a concrete upper bound. In particular, the choice of the intervals may depend on the (possibly unknown) functional variation.

\subsection{Convex Functions} \label{sec:convex}
For convex functions, we choose the meta-algorithm of \citet{Improved:Strongly:Adaptive} and take the online gradient descent as its subroutine. The following theorem regarding the adaptive regret can be obtained from that paper.
\begin{thm} \label{thm:2}  Under Assumption~\ref{ass:1}, the meta-algorithm of \citet{Improved:Strongly:Adaptive} is strongly adaptive with
\[
\begin{split}
\SAReg(T,\tau) \leq \left(\frac{12 BG}{\sqrt{2}-1} +8 \sqrt{7 \log T + 5} \right) \sqrt{\tau}  = O(\sqrt{\tau \log T}  ).
 \end{split}
\]
\end{thm}
From Theorems~\ref{thm:1} and \ref{thm:2}, we derive the following bound for the dynamic regret.
\begin{cor} \label{cor:convex} Under Assumption~\ref{ass:1}, the meta-algorithm of \citet{Improved:Strongly:Adaptive}  satisfies
\[
\begin{split}
\DReg(\w_1^*,\ldots,\w_T^*)\leq & \max\left\{\begin{split}
&  (c +9 \sqrt{7 \log T + 5})  \sqrt{T} \\
&  \frac{(c +8 \sqrt{5} )  T^{2/3} V_T^{1/3} }{\log^{1/6} T}    + 24 T^{2/3} V_T^{1/3} \log^{1/3} T
\end{split} \right.\\
= & O \left( \max\left\{  \sqrt{T \log T} , T^{2/3} V_T^{1/3} \log^{1/3} T  \right\} \right)
\end{split}
\]
where $c=12 BG/(\sqrt{2}-1)$.
\end{cor}
According to Theorem 2 of \citet{Non-Stationary}, we know that the minimax dynamic regret of convex functions is $O(T^{2/3} V_T^{1/3})$. Thus, our upper bound is minimax optimal up to a polylogarithmic factor. Although the restarted online gradient descent  of  \citet{Non-Stationary} achieves a dynamic regret of $O(T^{2/3}V_T^{1/3})$, it requires to know an upper bound of the functional variation $V_T$. In contrast, the meta-algorithm of  \citet{Improved:Strongly:Adaptive} does not need any prior knowledge of $V_T$. We note that the meta-algorithm of  \citet{Adaptive:ICML:15} can also be used here, and its dynamic regret  is on the order of $\max\left\{  \sqrt{T} \log T, T^{2/3} V_T^{1/3} \log^{2/3} T  \right\}$.
\subsection{Exponentially Concave Functions}
We proceed to consider exp-concave functions, defined in Definition~\ref{def:exp}. Exponential concavity is stronger than convexity but weaker than strong convexity. It can be used to model many popular losses used in machine learning, such as the square loss in regression, logistic loss in classification and negative logarithm loss in portfolio management~\citep{Sto:Exp:Con}.

For exp-concave functions, we choose Algorithm~\ref{alg:1} in this paper, and take the online Newton step as its subroutine. Based on Theorems~\ref{thm:adaptive} and \ref{thm:1}, we derive the dynamic regret of the proposed algorithm.
\begin{cor} \label{cor:exp} Let $K=\lceil T^{1/\gamma}\rceil$, where $\gamma>1$ is a small constant. Suppose Assumption~\ref{ass:1} holds, $\Omega \subset \R^d$, and all the functions are $\alpha$-exp-concave. Algorithm~\ref{alg:1}, with online Newton step as its subroutine, is strongly adaptive with
\[
\begin{split}
\SAReg(T,\tau)\leq& \left(\frac{(5d+1) (\gamma+1) + 2}{\alpha} + 5d (\gamma+1) GB \right) \log T\\
=&O\left(\gamma d \log T\right)=O\left( d \log T\right)
\end{split}
\]
and its dynamic regret satisfies
\[
\begin{split}
\DReg(\w_1^*,\ldots,\w_T^*)
\leq & \left(\frac{(5d+1) (\gamma+1) + 2}{\alpha} + 5d (\gamma+1) GB +2\right)  \max\left\{  \log T, \sqrt{T V_T \log T}  \right\} \\
= & O \left(d \cdot \max\left\{  \log T, \sqrt{T V_T \log T}  \right\} \right).
\end{split}
\]
\end{cor}
To the best of our knowledge, this is the \emph{first} dynamic regret that exploits exponential concavity. Furthermore, according to the minimax dynamic regret of strongly convex functions \citep{Non-Stationary}, our upper bound is  minimax optimal, up to a polylogarithmic factor.
\subsection{Strongly Convex Functions} \label{sec:strongly}
Finally, we study strongly convex functions. According to Lemma~\ref{lem:strongly}, we know that strongly convex functions with bounded gradients are also exp-concave. Thus, Corollary~\ref{cor:exp} can be directly applied to strongly convex functions, and yields a dynamic regret of $O(d \sqrt{T V_T \log T})$.  However, the upper bound depends on the dimensionality $d$. To address this limitation, we use online gradient descent as the subroutine in Algorithm~\ref{alg:1}.

From Theorems~\ref{thm:strong:convex} and \ref{thm:1}, we have the following theorem, in which both the adaptive and dynamic regrets are independent from $d$.
\begin{cor} \label{cor:strong:convex} Let $K=\lceil T^{1/\gamma}\rceil$, where $\gamma>1$ is a small constant. Suppose Assumption~\ref{ass:1} holds, and  all the functions are $\lambda$-strongly convex.  Algorithm~\ref{alg:1}, with online gradient descent as its subroutine, is strongly adaptive with
\[
\begin{split}
\SAReg(T,\tau) \leq  \frac{G^2}{2\lambda} \big(\gamma+1+ (3 \gamma+7)  \log T \big )= O\left(\gamma \log T\right)=O\left( \log T\right)
\end{split}
\]
and its dynamic regret satisfies
\[
\begin{split}
 \DReg(\w_1^*,\ldots,\w_T^*)\leq &\max\left\{\begin{split}
&  \frac{\gamma G^2 }{\lambda} + \left(\frac{5 \gamma G^2}{\lambda } +2 \right) \log T \\
&  \frac{\gamma G^2 }{\lambda} \sqrt{\frac{T V_T}{\log T} }+ \left(\frac{5 \gamma G^2}{\lambda } +2 \right)\sqrt{T V_T \log T}
\end{split} \right.\\
= & O \left( \max\left\{  \log T, \sqrt{T V_T \log T}  \right\} \right).
\end{split}
\]
\end{cor}
According to Theorem 4 of \citet{Non-Stationary}, the minimax dynamic regret of strongly convex functions is $O(\sqrt{T V_T})$, which implies our upper bound is almost minimax optimal. By comparison, the restarted online gradient descent  of  \citet{Non-Stationary} has a dynamic regret of $O(\log T \sqrt{T V_T})$, but it requires to know an upper bound of $V_T$.

\section{Analysis}
We here present the proof of main theorems.

\subsection{Proof of Theorem~\ref{thm:adaptive}}
From the second part of Lemma~\ref{lem:ending}, we know that there exist $m$ segments
\[
I_j = [t_j, e^{t_j}-1], \ j \in [m]
\]
with $m \leq \lceil\log_K (s-r+1)\rceil +1$, such that
\[
 t_1=r, \ e^{t_j}=t_{j+1}, \  j \in [m-1], \textrm{ and } e^{t_m} > s.
\]
Furthermore, the expert $E^{t_j}$ is alive during the period $[t_j, e^{t_j}-1]$.

Using Claim 3.1 of \citet{Hazan:2009:ELA}, we have
\[
\begin{split}
\sum_{t = t_j}^{e^{t_j}-1} f_t(\w_t) - f_t(\w^{t_j}_t) \leq  \frac{1}{\alpha}\left(\log t_j + 2\sum_{t = t_j + 1}^{e^{t_j}-1}\frac{1}{t}\right), \ \forall j \in [m-1]
\end{split}
\]
where $\w^{t_j}_{t_j}, \ldots, \w^{t_j}_{e^{t_j}-1}$ is the sequence of solutions generated by the expert $E^{t_j}$. Similarly, for the last segment, we have
\[
\begin{split}
\sum_{t = t_{m}}^{s} f_t(\w_t) - f_t(\w^{t_{m}}_t) \leq \frac{1}{\alpha}\left(\log  t_{m} + 2\sum_{t =  t_{m} + 1}^{s}\frac{1}{t}\right).
\end{split}
\]

By adding things together, we have
\begin{equation} \label{eqn:bound-1}
\begin{split}
& \sum_{j=1}^{m-1}   \left(\sum_{t = t_j}^{e^{t_j}-1} f_t(\w_t) - f_t(\w^{t_j}_t) \right)  + \sum_{t = t_{m}}^{s} f_t(\w_t) - f_t(\w^{t_{m}}_t)  \\
 \leq & \frac{1}{\alpha}\sum_{j=1}^m \log t_j  + \frac{2}{\alpha} \sum_{t=r+1}^s \frac{1}{t}  \leq  \frac{m + 2}{\alpha} \log T .
 \end{split}
\end{equation}
According to the property of online Newton step \citep[Theorem 2]{ML:Hazan:2007}, we have, for any $\w \in \Omega$,
\begin{equation}  \label{eqn:bound-2}
\sum_{t = t_j}^{e^{t_j}-1} f_t(\w^{t_j}_t) - f_t(\w) \leq 5d \left(\frac{1}{\alpha} +GB \right)\log T, \ \forall j \in [m-1]
\end{equation}
and
\begin{equation}  \label{eqn:bound-3}
\sum_{t = t_{m}}^{s}  f_t(\w^{t_m}_t) - f_t(\w) \leq 5d \left(\frac{1}{\alpha} +GB \right)\log T.
\end{equation}

Combining (\ref{eqn:bound-1}), (\ref{eqn:bound-2}), and (\ref{eqn:bound-3}), we have,
\[
\begin{split}
 \sum_{t=r}^s f_t(\w_t) -  \sum_{t=r}^s f_t(\w) \leq \left(\frac{(5d+1) m + 2}{\alpha} + 5d mGB \right) \log T
\end{split}
\]
for any $\w \in \Omega$.

\subsection{Proof of Theorem~\ref{thm:strong:convex}}
Lemma~\ref{lem:strongly} implies that all the $\lambda$-strongly convex functions are also $\frac{\lambda}{G^2}$-exp-concave. As a result, we can reuse the proof of Theorem~\ref{thm:adaptive}. Specifically, (\ref{eqn:bound-1}) with $\alpha=\frac{\lambda}{G^2}$ becomes
\begin{equation} \label{eqn:strong:convex:1}
\begin{split}
\sum_{j=1}^{m-1}   \left(\sum_{t = t_j}^{e^{t_j}-1} f_t(\w_t) - f_t(\w^{t_j}_t) \right)  + \sum_{t = t_{m}}^{s} f_t(\w_t) - f_t(\w^{t_{m}}_t)    \leq  \frac{(m + 2)G^2}{\lambda} \log T .
 \end{split}
\end{equation}
According to the property of online gradient descent \citep[Theorem 1]{ML:Hazan:2007}, we have, for any $\w \in \Omega$,
\begin{equation}  \label{eqn:strong:convex:2}
\sum_{t = t_j}^{e^{t_j}-1} f_t(\w^{t_j}_t) - f_t(\w) \leq \frac{G^2}{2\lambda} (1+\log T), \ \forall j \in [m-1]
\end{equation}
and
\begin{equation}  \label{eqn:strong:convex:3}
\sum_{t = t_{m}}^{s}  f_t(\w^{t_m}_t) - f_t(\w) \leq \frac{G^2}{2\lambda} (1+\log T).
\end{equation}
Combining (\ref{eqn:strong:convex:1}), (\ref{eqn:strong:convex:2}), and (\ref{eqn:strong:convex:3}), we have,
\[
\begin{split}
 \sum_{t=r}^s f_t(\w_t) -  \sum_{t=r}^s f_t(\w) \leq  \frac{G^2}{2\lambda} \big(m+ (3 m +4)  \log T \big)
\end{split}
\]
for any $\w \in \Omega$.

\subsection{Proof of Theorem~\ref{thm:1}}
First, we upper bound the dynamic regret in the following way
\begin{equation} \label{eqn:thm1:1}
\begin{split}
&\DReg(\w_1^*,\ldots,\w_T^*) \\
= &  \sum_{i=1}^k \left(\sum_{t=s_i}^{q_i} f_t(\w_t)  - \sum_{t=s_i}^{q_i} \min_{\w \in \Omega} f_t(\w) \right)\\
= & \sum_{i=1}^k \left( \underbrace{\sum_{t=s_i}^{q_i} f_t(\w_t) - \min_{\w \in \Omega} \sum_{t=s_i}^{q_i} f_t(\w)}_{:=a_i} +\underbrace{\min_{\w \in \Omega} \sum_{t=s_i}^{q_i} f_t(\w)- \sum_{t=s_i}^{q_i} \min_{\w \in \Omega} f_t(\w)}_{:=b_i} \right).
\end{split}
\end{equation}

From the definition of strongly adaptive regret, we can upper bound $a_i$ by
\[
\sum_{t=s_i}^{q_i} f_t(\w_t) - \min_{\w \in \Omega} \sum_{t=s_i}^{q_i} f_t(\w) \leq \SAReg(T,|\I_i|).
\]
To upper bound $b_i$, we follow the analysis of Proposition 2 of \citet{Non-Stationary}:
\begin{equation} \label{eqn:thm1:bt:1}
\begin{split}
&\min_{\w \in \Omega} \sum_{t=s_i}^{q_i} f_t(\w)- \sum_{t=s_i}^{q_i} \min_{\w \in \Omega} f_t(\w)= \min_{\w \in \Omega} \sum_{t=s_i}^{q_i} f_t(\w)- \sum_{t=s_i}^{q_i}  f_t(\w_t^*) \\
\leq & \sum_{t=s_i}^{q_i} f_t(\w_{s_i}^*)- \sum_{t=s_i}^{q_i}  f_t(\w_t^*) \leq |\I_i| \cdot \max_{t \in [s_i,q_i]} \left( f_t(\w_{s_i}^*)- f_t(\w_t^*) \right).
\end{split}
\end{equation}
Furthermore, for any $t \in [s_i,q_i]$, we have
\begin{equation} \label{eqn:thm1:bt:2}
\begin{split}
 &f_t(\w_{s_i}^*)- f_t(\w_t^*)  =  f_t(\w_{s_i}^*)- f_{s_i}(\w_{s_i}^*) + f_{s_i}(\w_{s_i}^*)  -  f_t(\w_t^*) \\
\leq & f_t(\w_{s_i}^*)- f_{s_i}(\w_{s_i}^*) + f_{s_i}(\w_t^*)  -  f_t(\w_t^*) \leq  2 V_T(i).
\end{split}
\end{equation}
Combining (\ref{eqn:thm1:bt:1}) with (\ref{eqn:thm1:bt:2}), we have
\[
\min_{\w \in \Omega} \sum_{t=s_i}^{q_i} f_t(\w)- \sum_{t=s_i}^{q_i} \min_{\w \in \Omega} f_t(\w)  \leq 2  |\I_i| \cdot V_T(i).
\]

Substituting the upper bounds of $a_i$ and $b_i$ into (\ref{eqn:thm1:1}), we arrive at
\[
\begin{split}
\DReg(\w_1^*,\ldots,\w_T^*)\leq   \sum_{i=1}^k \left(  \SAReg(T,|\I_i|)+ 2  |\I_i| \cdot V_T(i) \right).
\end{split}
\]
Since the above inequality holds for any partition of $[1,T]$, we can take minimization to get a tight bound.

\subsection{Proof of Corollary~\ref{cor:convex}}
To simplify the upper bound in Theorem~\ref{thm:1}, we restrict to intervals of the same length $\tau$, and in this case $k=T/\tau$. Then, we have
\[
\begin{split}
  \DReg(\w_1^*,\ldots,\w_T^*) \leq &  \min_{1 \leq \tau \leq T} \sum_{i=1}^k \big( \SAReg(T,\tau) + 2 \tau  V_T(i) \big)\\
= & \min_{1 \leq \tau \leq T}  \left(  \frac{ \SAReg(T,\tau) T}{\tau} + 2 \tau  \sum_{i=1}^k V_T(i) \right)\\
\leq & \min_{1 \leq \tau \leq T}  \left( \frac{ \SAReg(T,\tau) T}{\tau} + 2 \tau V_T \right).
\end{split}
\]
Combining with Theorem~\ref{thm:2}, we have
\[
\begin{split}
  \DReg(\w_1^*,\ldots,\w_T^*)\leq  \min_{1 \leq \tau \leq T}  \left( \frac{(c +8 \sqrt{7 \log T + 5})  T }{\sqrt{\tau}} + 2 \tau V_T \right).
\end{split}
\]
where $c=12 BG/(\sqrt{2}-1)$.

In the following, we consider two cases. If $V_T \geq \sqrt{\log T/T}$, we choose
\[
\tau = \left( \frac{T \sqrt{\log T}}{V_T} \right)^{2/3} \leq T
\]
and have
\[
\begin{split}
\DReg(\w_1^*,\ldots,\w_T^*)  \leq  &  \frac{(c +8 \sqrt{7 \log T + 5} )  T^{2/3} V_T^{1/3} }{\log^{1/6} T}    + 2 T^{2/3} V_T^{1/3} \log^{1/3} T\\
 \leq & \frac{(c +8 \sqrt{5} )  T^{2/3} V_T^{1/3} }{\log^{1/6} T}    + (2+ 8\sqrt{7}) T^{2/3} V_T^{1/3} \log^{1/3} T.
\end{split}
\]
Otherwise, we choose  $\tau=T$, and have
\[
\begin{split}
 \DReg(\w_1^*,\ldots,\w_T^*)  \leq &(c +8 \sqrt{7 \log T + 5})  \sqrt{T}  + 2 T V_T \\
  \leq & (c +8 \sqrt{7 \log T + 5})  \sqrt{T}  + 2 T \sqrt{\frac{\log T}{T}} \\
\leq & (c +9 \sqrt{7 \log T + 5})  \sqrt{T}.
 \end{split}
\]

In summary, we have
\[
\begin{split}
\DReg(\w_1^*,\ldots,\w_T^*)\leq & \max\left\{\begin{split}
&  (c +9 \sqrt{7 \log T + 5})  \sqrt{T} \\
&  \frac{(c +8 \sqrt{5} )  T^{2/3} V_T^{1/3} }{\log^{1/6} T}    + 24 T^{2/3} V_T^{1/3} \log^{1/3} T
\end{split} \right.\\
= & O \left( \max\left\{  \sqrt{T \log T} , T^{2/3} V_T^{1/3} \log^{1/3} T  \right\} \right).
\end{split}
\]

\subsection{Proof of Corollary~\ref{cor:exp}}
The first part of Corollary~\ref{cor:exp} is a direct consequence of Theorem~\ref{thm:adaptive} by setting $K=\lceil T^{1/\gamma} \rceil$.

Now, we prove the second part. Following similar analysis of Corollary~\ref{cor:convex}, we have
\[
\begin{split}
 \DReg(\w_1^*,\ldots,\w_T^*) \leq  \min_{1 \leq \tau \leq T}  \left\{ \left(\frac{(5d+1) (\gamma+1) + 2}{\alpha} + 5d (\gamma+1) GB \right) \frac{ T \log T }{\tau} + 2 \tau V_T \right\}.
\end{split}
\]
Then, we consider two cases. If $V_T \geq \log T/T$, we choose
\[
\tau = \sqrt{ \frac{T \log T }{V_T}} \leq T
\]
and have
\[
\begin{split}
\DReg(\w_1^*,\ldots,\w_T^*)  \leq   \left(\frac{(5d+1) (\gamma+1) + 2}{\alpha} + 5d (\gamma+1) GB +2\right)\sqrt{T V_T \log T} .
\end{split}
\]
Otherwise, we choose  $\tau=T$, and have
\[
\begin{split}
\DReg(\w_1^*,\ldots,\w_T^*) \leq & \left(\frac{(5d+1) (\gamma+1) + 2}{\alpha} + 5d (\gamma+1) GB \right)  \log T + 2 T V_T \\
\leq &\left(\frac{(5d+1) (\gamma+1) + 2}{\alpha} + 5d (\gamma+1) GB \right)  \log T + 2 T \frac{\log T}{T}\\
= &\left(\frac{(5d+1) (\gamma+1) + 2}{\alpha} + 5d (\gamma+1) GB +2\right) \log T .
\end{split}
\]

In summary, we have
\[
\begin{split}
 \DReg(\w_1^*,\ldots,\w_T^*)
\leq & \left(\frac{(5d+1) (\gamma+1) + 2}{\alpha} + 5d (\gamma+1) GB +2\right)
  \max\left\{  \log T, \sqrt{T V_T \log T}  \right\} \\
= & O \left(d \cdot \max\left\{  \log T, \sqrt{T V_T \log T}  \right\} \right).
\end{split}
\]

\subsection{Proof of Corollary~\ref{cor:strong:convex}}
The first part of Corollary~\ref{cor:strong:convex} is a direct consequence of Theorem~\ref{thm:strong:convex} by setting $K=\lceil T^{1/\gamma}\rceil$.

The proof of the second part is similar to that of Corollary~\ref{cor:exp}. First, we have
\[
\begin{split}
 \DReg(\w_1^*,\ldots,\w_T^*) \leq & \min_{1 \leq \tau \leq T}  \left\{ \frac{G^2}{2\lambda} \big(\gamma+1+ (3 \gamma+7)  \log T \big) \frac{T}{\tau} + 2 \tau V_T \right\}\\
 \leq &  \min_{1 \leq \tau \leq T}  \left\{  \frac{( \gamma+ 5 \gamma  \log T )G^2 T}{\lambda \tau} + 2 \tau V_T \right\}
\end{split}
\]
where the last inequality is due to the condition $\gamma >1$.

Then, we consider two cases. If $V_T \geq \log T/T$, we choose
\[
\tau = \sqrt{ \frac{T \log T }{V_T}} \leq T
\]
and have
\[
\begin{split}
\DReg(\w_1^*,\ldots,\w_T^*)  \leq & \frac{\gamma G^2 }{\lambda} \sqrt{\frac{T V_T}{\log T} }+ \frac{5 \gamma G^2}{\lambda } \sqrt{T V_T \log T} +  2 \sqrt{T V_T \log T} \\
=&\frac{\gamma G^2 }{\lambda} \sqrt{\frac{T V_T}{\log T} }+ \left(\frac{5 \gamma G^2}{\lambda } +2 \right)\sqrt{T V_T \log T}  .
\end{split}
\]
Otherwise, we choose  $\tau=T$, and have
\[
\begin{split}
\DReg(\w_1^*,\ldots,\w_T^*) \leq & \frac{( \gamma+ 5 \gamma  \log T )G^2}{\lambda } + 2 T V_T \\
\leq  &  \frac{( \gamma+ 5 \gamma  \log T )G^2}{\lambda } +2 T \frac{\log T}{T} \\
=& \frac{\gamma G^2 }{\lambda} + \left(\frac{5 \gamma G^2}{\lambda } +2 \right) \log T.
\end{split}
\]

In summary, we have
\[
\begin{split}
 \DReg(\w_1^*,\ldots,\w_T^*)
\leq &\max\left\{\begin{split}
&  \frac{\gamma G^2 }{\lambda} + \left(\frac{5 \gamma G^2}{\lambda } +2 \right) \log T \\
&  \frac{\gamma G^2 }{\lambda} \sqrt{\frac{T V_T}{\log T} }+ \left(\frac{5 \gamma G^2}{\lambda } +2 \right)\sqrt{T V_T \log T}
\end{split} \right.\\
= & O \left( \max\left\{  \log T, \sqrt{T V_T \log T}  \right\} \right).
\end{split}
\]

\section{Conclusions and Future Work}
In this paper, we demonstrate that the dynamic regret can be upper bounded by the adaptive regret and the functional variation, which implies strongly adaptive algorithms are automatically equipped with tight dynamic regret bounds. As a result, we are able to derive dynamic regret bounds for convex functions, exp-concave functions, and strongly convex functions. Moreover, we provide a unified approach for minimizing the adaptive regret of exp-concave functions, as well as strongly convex functions.

The adaptive-to-dynamic conversion  leads to a series of dynamic regret bounds in terms of the functional variation.  As we mentioned before, dynamic regret can also be upper bounded by other regularities such as the path-length. It is interesting to investigate whether those kinds of upper bounds can also be established for strongly adaptive algorithms.

\appendix
\section{Proof of Lemma~\ref{lem:ending}}
We first prove the first part of Lemma~\ref{lem:ending}.  Let $k= \lfloor \log_K t \rfloor$. Then, integer $t$ can be represented in the base-$K$ number system as
\[
t = \sum_{j=0}^k \beta_j K^j .
\]
From the definition of base-$K$ ending time, integers that are no larger than $t$ and alive at $t$ are
\[
\left\{\begin{split}
&1*K^0+\sum_{j=1}^k \beta_j K^j, \  2*K^0+ \sum_{j=1}^k \beta_j K^j, \ \ldots, \ \beta_0*K^0 + \sum_{j=1}^k \beta_j K^j \\
&1*K^1+\sum_{j=2}^k \beta_j K^j, \  2*K^1+ \sum_{j=2}^k \beta_j K^j, \ \ldots, \ \beta_1*K^1 + \sum_{j=2}^k \beta_j K^j \\
&\ldots  \\
&1*K^{k-1}+\beta_k K^k, \  1*K^{k-1}+  \beta_k K^k, \ \ldots, \ \beta_{k-1}*K^{k-1} +\beta_k K^k \\
&1 *K^k, \  2 * K^k, \  \ldots, \ \beta_k K^k \\
\end{split} \right\}.
\]
The total number of alive integers are upper bounded by
\[
\sum_{i=0}^k \beta_i \leq (k+1) (K-1)= (\lfloor \log_K t \rfloor+1) (K-1).
\]

We proceed to prove the second part of Lemma~\ref{lem:ending}.  Let $k= \lfloor \log_K r \rfloor$, and the representation of $r$ in the base-$K$ number system be
\[
r= \sum_{j=0}^k \beta_j K^j.
\]
We generate a sequence of segments as
\[
\begin{split}
 I_1& = [t_1, e^{t_1}-1] =\left[\sum_{j=0}^k \beta_j K^j, (\beta_1+1) K^1+\sum_{j=2}^k \beta_j K^j - 1\right], \\
 I_2& = [t_2, e^{t_2}-1] = \left[(\beta_1+1) K^1+\sum_{j=2}^k \beta_j K^j, (\beta_2+1) K^2+\sum_{j=3}^k \beta_j K^j - 1\right], \\
  I_3& = [t_3, e^{t_3}-1] = \left[(\beta_2+1) K^2+\sum_{j=3}^k \beta_j K^j, (\beta_3+1) K^3+\sum_{j=4}^k \beta_j K^j - 1\right], \\
& \ldots \\
 I_k& = [t_k, e^{t_k}-1] = \left[(\beta_{k-1}+1)K^{k-1} +\beta_k K^k, (\beta_k+1) K^k - 1\right], \\
 I_{k+1}& = [t_{k+1}, e^{t_{k+1}}-1] = \left[(\beta_k+1) K^k, K^{k+1}-1\right], \\
 I_{k+2}& = [t_{k+2}, e^{t_{k+2}}-1] = \left[K^{k+1}, K^{k+2}-1\right], \\
& \ldots
\end{split}
\]
until $s$ is covered. It is easy to verify that
\[
t_{m+1} > t_{m} + K^{m-1} -1.
\]
Thus, $s$ will be covered by the first $m$ intervals as long as
\[
t_{m} + K^{m-1} -1  \geq s.
\]
A sufficient condition is
\[
r+ K^{m-1} -1  \geq s
\]
which is satisfied when
\[
m=\lceil\log_K (s-r+1)\rceil +1.
\]

\section{Proof of Lemma~\ref{lem:strongly}}
The gradient of $\exp(-\alpha f(\w))$ is
\[
\nabla \exp(-\alpha f(\w)) =  \exp(-\alpha f(\w))  {-}\alpha \nabla f(\w) = {-} \alpha  \exp({-}\alpha f(\w)) \nabla f(\w).
\]
and the Hessian is
\[
\begin{split}
\nabla^2 \exp(-\alpha f(\w)) = & {-} \alpha  \exp(-\alpha f(\w)) {-} \alpha  \nabla f(\w) \nabla^{\top} f(\w) -\alpha \exp(-\alpha f(\w)) \nabla^2 f(\w) \\
= & \alpha  \exp(-\alpha f(\w)) \left(\alpha  \nabla f(\w) \nabla^{\top} f(\w)  -  \nabla^2 f(\w)  \right).
\end{split}
\]
Thus, $f(\cdot)$ is $\alpha$-exp-concave if
\[
\alpha  \nabla f(\w) \nabla^{\top} f(\w)  \preceq  \nabla^2 f(\w).
\]
We complete the proof by noticing
\[
\frac{\lambda}{G^2}  \nabla f(\w) \nabla^{\top} f(\w) \preceq    \lambda I \preceq \nabla^2 f(\w).
\]

\section{Proof of Theorem~\ref{thm:2}}
As pointed out by \citet{Adaptive:ICML:15}, the static regret of online gradient descent \citep{zinkevich-2003-online} over any interval of length $\tau$ is upper bounded by $3BG\sqrt{\tau}$. Combining this fact with Theorem 2 of \citet{Improved:Strongly:Adaptive}, we get Theorem~\ref{thm:2} in this paper.
\bibliography{E:/MyPaper/ref}

\begin{thebibliography}{27}
\providecommand{\natexlab}[1]{#1}
\providecommand{\url}[1]{\texttt{#1}}
\expandafter\ifx\csname urlstyle\endcsname\relax
  \providecommand{\doi}[1]{doi: #1}\else
  \providecommand{\doi}{doi: \begingroup \urlstyle{rm}\Url}\fi

\bibitem[Abernethy et~al.(2009)Abernethy, Agarwal, Bartlett, and
  Rakhlin]{Minimax:Regret}
Jacob Abernethy, Alekh Agarwal, Peter~L. Bartlett, and Alexander Rakhlin.
\newblock A stochastic view of optimal regret through minimax duality.
\newblock In \emph{Proceedings of the 22nd Annual Conference on Learning
  Theory}, 2009.

\bibitem[Adamskiy et~al.(2012)Adamskiy, Koolen, Chernov, and
  Vovk]{Adamskiy2012}
Dmitry Adamskiy, Wouter~M. Koolen, Alexey Chernov, and Vladimir Vovk.
\newblock A closer look at adaptive regret.
\newblock In \emph{Proceedings of the 23rd International Conference on
  Algorithmic Learning Theory}, pages 290--304, 2012.

\bibitem[Besbes et~al.(2015)Besbes, Gur, and Zeevi]{Non-Stationary}
Omar Besbes, Yonatan Gur, and Assaf Zeevi.
\newblock Non-stationary stochastic optimization.
\newblock \emph{Operations Research}, 63\penalty0 (5):\penalty0 1227--1244,
  2015.

\bibitem[Cesa-Bianchi and Lugosi(2006)]{bianchi-2006-prediction}
Nicol\`{o} Cesa-Bianchi and G{\'a}bor Lugosi.
\newblock \emph{Prediction, Learning, and Games}.
\newblock Cambridge University Press, 2006.

\bibitem[Cesa-bianchi et~al.(2012)Cesa-bianchi, Gaillard, Lugosi, and
  Stoltz]{Fixed:Share:NIPS12}
Nicol\`{o} Cesa-bianchi, Pierre Gaillard, Gabor Lugosi, and Gilles Stoltz.
\newblock Mirror descent meets fixed share (and feels no regret).
\newblock In \emph{Advances in Neural Information Processing Systems 25}, pages
  980--988, 2012.

\bibitem[Daniely et~al.(2015)Daniely, Gonen, and
  Shalev-Shwartz]{Adaptive:ICML:15}
Amit Daniely, Alon Gonen, and Shai Shalev-Shwartz.
\newblock Strongly adaptive online learning.
\newblock In \emph{Proceedings of the 32nd International Conference on Machine
  Learning}, pages 1405--1411, 2015.

\bibitem[Gy\"{o}rgy et~al.(2012)Gy\"{o}rgy, Linder, and
  Lugosi]{Track_Large_Expert}
Andr\'{a}s Gy\"{o}rgy, Tam\'{a}s Linder, and G\'{a}bor Lugosi.
\newblock Efficient tracking of large classes of experts.
\newblock \emph{IEEE Transactions on Information Theory}, 58\penalty0
  (11):\penalty0 6709--6725, 2012.

\bibitem[Hall and Willett(2013)]{Dynamic:ICML:13}
Eric~C. Hall and Rebecca~M. Willett.
\newblock Dynamical models and tracking regret in online convex programming.
\newblock In \emph{Proceedings of the 30th International Conference on Machine
  Learning}, pages 579--587, 2013.

\bibitem[Hazan and Seshadhri(2007)]{Adaptive:Hazan}
Elad Hazan and C.~Seshadhri.
\newblock Adaptive algorithms for online decision problems.
\newblock \emph{Electronic Colloquium on Computational Complexity}, 88, 2007.

\bibitem[Hazan and Seshadhri(2009)]{Hazan:2009:ELA}
Elad Hazan and C.~Seshadhri.
\newblock Efficient learning algorithms for changing environments.
\newblock In \emph{Proceedings of the 26th Annual International Conference on
  Machine Learning}, pages 393--400, 2009.

\bibitem[Hazan et~al.(2007)Hazan, Agarwal, and Kale]{ML:Hazan:2007}
Elad Hazan, Amit Agarwal, and Satyen Kale.
\newblock Logarithmic regret algorithms for online convex optimization.
\newblock \emph{Machine Learning}, 69\penalty0 (2-3):\penalty0 169--192, 2007.

\bibitem[Herbster and Warmuth(1998)]{Herbster1998}
Mark Herbster and Manfred~K. Warmuth.
\newblock Tracking the best expert.
\newblock \emph{Machine Learning}, 32\penalty0 (2):\penalty0 151--178, 1998.

\bibitem[Herbster and Warmuth(2001)]{Herbster:2001:TBL}
Mark Herbster and Manfred~K. Warmuth.
\newblock Tracking the best linear predictor.
\newblock \emph{Journal of Machine Learning Research}, 1:\penalty0 281--309,
  2001.

\bibitem[Jadbabaie et~al.(2015)Jadbabaie, Rakhlin, Shahrampour, and
  Sridharan]{Oinline:Dynamic:Comp}
Ali Jadbabaie, Alexander Rakhlin, Shahin Shahrampour, and Karthik Sridharan.
\newblock Online optimization : Competing with dynamic comparators.
\newblock In \emph{Proceedings of the 18th International Conference on
  Artificial Intelligence and Statistics}, 2015.

\bibitem[Jun et~al.(2017)Jun, Orabona, Wright, and
  Willett]{Improved:Strongly:Adaptive}
Kwang-Sung Jun, Francesco Orabona, Stephen Wright, and Rebecca Willett.
\newblock {Improved Strongly Adaptive Online Learning using Coin Betting}.
\newblock In \emph{Proceedings of the 20th International Conference on
  Artificial Intelligence and Statistics}, pages 943--951, 2017.

\bibitem[Koren(2013)]{Sto:Exp:Con}
Tomer Koren.
\newblock Open problem: Fast stochastic exp-concave optimization.
\newblock In \emph{Proceedings of the 26th Annual Conference on Learning
  Theory}, pages 1073--1075, 2013.

\bibitem[Langford et~al.(2009)Langford, Li, and Zhang]{Sparse:Online}
John Langford, Lihong Li, and Tong Zhang.
\newblock Sparse online learning via truncated gradient.
\newblock In \emph{Advances in Neural Information Processing Systems 21}, pages
  905--912, 2009.

\bibitem[Littlestone and Warmuth(1994)]{LITTLESTONE1994212}
Nick Littlestone and Manfred~K. Warmuth.
\newblock The weighted majority algorithm.
\newblock \emph{Information and Computation}, 108\penalty0 (2):\penalty0
  212--261, 1994.

\bibitem[Mokhtari et~al.(2016)Mokhtari, Shahrampour, Jadbabaie, and
  Ribeiro]{Dynamic:Strongly}
Aryan Mokhtari, Shahin Shahrampour, Ali Jadbabaie, and Alejandro Ribeiro.
\newblock Online optimization in dynamic environments: Improved regret rates
  for strongly convex problems.
\newblock In \emph{IEEE 55th Conference on Decision and Control}, pages
  7195--7201, 2016.

\bibitem[Shalev-Shwartz(2011)]{Online:suvery}
Shai Shalev-Shwartz.
\newblock Online learning and online convex optimization.
\newblock \emph{Foundations and Trends in Machine Learning}, 4\penalty0
  (2):\penalty0 107--194, 2011.

\bibitem[Shalev-Shwartz and Singer(2007)]{Shalev:Primal:Dual}
Shai Shalev-Shwartz and Yoram Singer.
\newblock A primal-dual perspective of online learning algorithms.
\newblock \emph{Machine Learning}, 69\penalty0 (2):\penalty0 115--142, 2007.

\bibitem[Shalev-Shwartz et~al.(2007)Shalev-Shwartz, Singer, and
  Srebro]{ICML_Pegasos}
Shai Shalev-Shwartz, Yoram Singer, and Nathan Srebro.
\newblock Pegasos: primal estimated sub-gradient solver for {SVM}.
\newblock In \emph{Proceedings of the 24th International Conference on Machine
  Learning}, pages 807--814, 2007.

\bibitem[Wang et~al.(2018)Wang, Zhao, and Zhang]{Adaptive:One:Gradient}
Guanghui Wang, Dakuan Zhao, and Lijun Zhang.
\newblock Minimizing adaptive regret with one gradient per iteration.
\newblock In \emph{Proceedings of the 27th International Joint Conference on
  Artificial Intelligence}, 2018.

\bibitem[Yang et~al.(2016)Yang, Zhang, Jin, and Yi]{Dynamic:2016}
Tianbao Yang, Lijun Zhang, Rong Jin, and Jinfeng Yi.
\newblock Tracking slowly moving clairvoyant: Optimal dynamic regret of online
  learning with true and noisy gradient.
\newblock In \emph{Proceedings of the 33rd International Conference on Machine
  Learning}, pages 449--457, 2016.

\bibitem[Zhang et~al.(2013)Zhang, Yi, Jin, Lin, and He]{ICML:13:Zhang:Sparse}
Lijun Zhang, Jinfeng Yi, Rong Jin, Ming Lin, and Xiaofei He.
\newblock Online kernel learning with a near optimal sparsity bound.
\newblock In \emph{Proceedings of the 30th International Conference on Machine
  Learning}, 2013.

\bibitem[Zhang et~al.(2017)Zhang, Yang, Yi, Jin, and
  Zhou]{Dynamic:Regret:Squared}
Lijun Zhang, Tianbao Yang, Jinfeng Yi, Rong Jin, and Zhi-Hua Zhou.
\newblock Improved dynamic regret for non-degenerate functions.
\newblock In \emph{Advances in Neural Information Processing Systems 30}, pages
  732--741, 2017.

\bibitem[Zinkevich(2003)]{zinkevich-2003-online}
Martin Zinkevich.
\newblock Online convex programming and generalized infinitesimal gradient
  ascent.
\newblock In \emph{Proceedings of the 20th International Conference on Machine
  Learning}, pages 928--936, 2003.

\end{thebibliography}

\end{document}